\documentclass{IOS-Book-Article}

\usepackage{mathptmx}
\usepackage{soul}\setuldepth{article}

\usepackage{xcolor}
\usepackage{booktabs}
\usepackage{amsmath} 
\usepackage{cleveref}
\usepackage{graphicx}
\usepackage{quoting}
\usepackage{subcaption}

\newcommand{\system}{HILL}
\newcommand{\samplemean}{\bar{x}}
\newcommand{\samplesd}{s}
\newcommand{\scenarioA}{CIFAR-10}
\newcommand{\scenarioB}{PAMAP2}

\renewenvironment{quoting}
  {\list{}{\leftmargin=7mm\rightmargin=0mm \parsep=0pt \topsep=1pt}\item[]}
  {\endlist}

\def\hb{\hbox to 11.5 cm{}}

\begin{document}

\pagestyle{headings}
\def\thepage{}
\begin{frontmatter}              

\title{Human in the Latent Loop (\system): Interactively Guiding Model Training Through Human Intuition}

\author[A]{\fnms{Daniel} \snm{Geißler}}%
\author[A]{\fnms{Lars} \snm{Krupp}}
\author[A]{\fnms{Vishal} \snm{Banwari}}
\author[A]{\fnms{David} \snm{Habusch}}
\author[A,B]{\fnms{Bo} \snm{Zhou}}
\author[A,B]{\fnms{Paul} \snm{Lukowicz}}
\author[A,B]{\fnms{Jakob} \snm{Karolus}}
\runningauthor{Geißler et al.}
\runningtitle{HILL: Interactively Guiding Model Training Through Human Intuition}
\address[A]{German Research Center for Artificial Intelligence (DFKI), Germany}
\address[A]{University of Kaiserslautern-Landau (RPTU), Germany}


\begin{abstract}
Latent space representations are critical for understanding and improving the behavior of machine learning models, yet they often remain obscure and intricate. 
Understanding and exploring the latent space has the potential to contribute valuable human intuition and expertise about respective domains.
In this work, we present \system, an interactive framework allowing users to incorporate human intuition into the model training by interactively reshaping latent space representations.
The modifications are infused into the model training loop via a novel approach inspired by knowledge distillation, treating the user's modifications as a teacher to guide the model in reshaping its intrinsic latent representation.
The process allows the model to converge more effectively and overcome inefficiencies, as well as provide beneficial insights to the user.
We evaluated \system{} in a user study tasking participants to train an optimal model, closely observing the employed strategies.
The results demonstrated that human-guided latent space modifications enhance model performance while maintaining generalization, yet also revealing the risks of including user biases.
Our work introduces a novel human-AI interaction paradigm that infuses human intuition into model training and critically examines the impact of human intervention on training strategies and potential biases.


\end{abstract}

\begin{keyword}
Latent Space, Interactive Learning, Human Intuition
\end{keyword}

\end{frontmatter}

\section{Introduction}

\begin{figure}[ht] 
    \centering
    \includegraphics[width=0.75\textwidth]{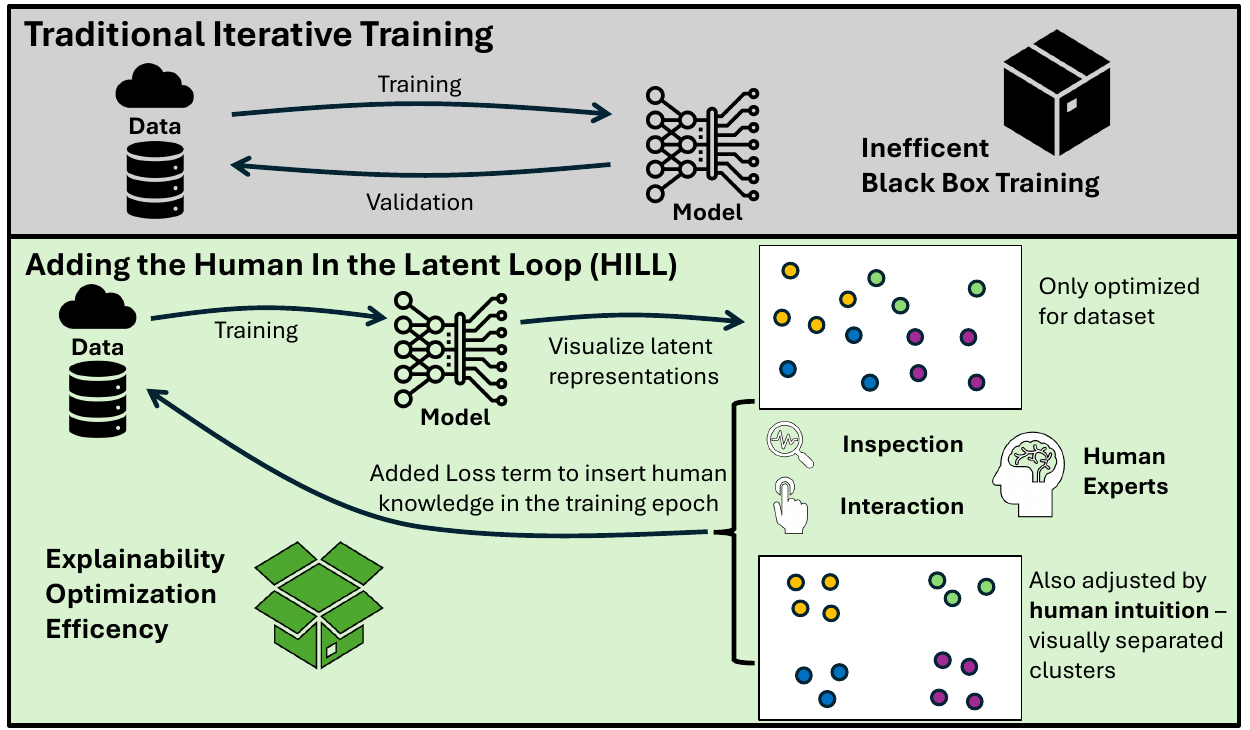} 
    \caption{Comparison of traditional iterative training, commonly resulting in a unsatisfying black-box training, with the approach of adding the Human In the Latent Loop (\system{}) to insert the knowledge of human experts towards explainability, optimization and efficiency.}
    \label{fig:teaser}
    \vspace{-0.75em}
\end{figure}

The interplay between layers in deep learning models forms latent representations that capture the inherent properties and relationships within data. Through a progressive learning process, each layer extracts and refines specific features, transforming raw input into higher-level abstractions that ultimately guide the model’s decisions \cite{Janiesch2021Machine}. This distributed learning approach enables deep models to encode complex patterns effectively~\cite{lecun2015deep}.
Despite the central role of latent representations in decision-making, their structure and behavior are often opaque and obscure to human interpretation, making it difficult to assess or influence how the model encodes the information internally \cite{Qamar2023Understanding,Li2021Interpretable}. 
While much research focused on optimizing models through loss functions and architecture design, little attention has been given to directly examining and influencing how data is represented within the latent spaces of intermediate layers during training \cite{Murugesan2019DeepCompare}.
Existing methods primarily focus on post hoc visualizations or indirect feature engineering, ignoring the potential of training-time, and interactive human feedback to enhance model performance and interpretability \cite{geissler2023latent,wei2022spaceediting}. Consequently, there is a need for approaches that allow for more profound insight into the internal dynamics of model learning and create opportunities for humans to guide and refine these processes \cite{gilpin2018explaining}.


In this work, we propose a novel human-AI interaction paradigm at training time called Human In the Latent Loop (\system), combining human intuition with latent space exploration and integrating it into the training loop (\Cref{fig:teaser}). 
By visualizing and allowing modification of latent spaces during model training, users can directly influence the structure of the latent space to reflect human intuition, making use of their intrinsic domain knowledge and human discernment.
Our approach relies on a knowledge distillation mechanism, where user-guided adjustments act as knowledge teacher to guide the model in refining its internal latent representation. As such, our approach does not manipulate the model nor the data 
but rather guides the model to refine the learning mechanism according to the user's input.
We evaluated \system{} in a user study tasking participants to find optimal models for two well-known machine learning (ML) datasets. The participants were supported by \system{} visualizing the latent space and iteratively refined it based on their intuition to improve the model. We recorded the employed strategies and acquired feedback from participants through post-hoc questionnaires and interviews. 
Our results show that participants used diverse strategies, such as increasing cluster compactness or maximizing class separation, leading to improved accuracy and faster convergence compared to baseline training.


Our contributions can be summarized as follows:
\vspace{-0.5em}
\begin{enumerate}
    \item We introduce \system, an interactive tool for latent space exploration that allows users to monitor and guide the model directly during training. \system{} is open-source and available on GitHub (\href{https://github.com/DFKIEI/HITL-ML}{https://github.com/DFKIEI/HITL-ML}).
    \item We develop a loss function insertion approach inspired by knowledge distillation, that integrates user modifications into the training process, enhancing the structural organization of latent spaces.
    \item We evaluate our framework in a user study, demonstrating its effectiveness while critically examining the aspects of human intuition and its potential manipulation.
\end{enumerate}

\section{Related Work}
\label{sec:RL}
Human-in-the-loop (HITL) ML is an emerging paradigm that incorporates human expertise into the model development process to enhance performance while addressing inherent limitations \cite{teso2019explanatory}.
By integrating human judgment and intuition, HITL systems provide adaptability and reliability in applications requiring specialized domain expertise \cite{stumpf2009interacting}.

HITL techniques have demonstrated success in various ML tasks, particularly in data preprocessing, annotation, and iterative labeling.
For instance, humans can help transform unstructured data into structured formats, ensuring the creation of high-quality training datasets \cite{xin2018accelerating}. 
Human-guided workflows also enable the creation of rules for data extraction, improving consistency and reducing computational burdens \cite{kulesza2015principles}. 
In health research, HITL approaches have been used to address tasks like subspace clustering and k-anonymization, where human intervention reduces algorithmic complexity while maintaining accuracy \cite{holzinger2016interactive}. 
Additionally, HITL systems in interactive ML frameworks empower users to iteratively refine model outputs, achieving improved performance in tasks such as classification, regression, and clustering \cite{fails2003interactive}.

Beyond data preparation, HITL has played a significant role in enhancing the interpretability of ML models. 
Deep reinforcement learning allows to generate human-readable explanations for model predictions in critical domains like medical diagnostics \cite{ribeiro2016should}. 
Similarly, concept-based methods enable humans to interact with interpretable latent spaces, facilitating actionable refinements and diagnostics for model behavior \cite{kim2018interpretability}.

Integrating human feedback into the training loop presents challenges. 
One prominent issue is the variability in user strategies leading to inconsistent modifications \cite{geissler2024strategies}. 
Thus, achieving uniform improvements across user interactions to design reproducible workflows is complex~\cite{guidotti2018survey}.
Additionally, cognitive biases inherent in human decision-making may inadvertently introduce inaccuracies, requiring careful handling of human-algorithm dynamics \cite{jakubik2022empirical}.

Consequently, HITL systems must compromise between leveraging human intuition and preserving the computational rigor of ML algorithms. 
Moreover, ensuring transparency and accountability in workflows is essential to maintain trust in the system \cite{lipton2018mythos}. 
Scalability issues further challenge real-time integration of user inputs~\cite{settles2009active}.

Recent advances in HITL research focus on active learning and real-time feedback mechanisms, enabling models to adapt more dynamically during training \cite{chai2020human}.
Central to this progress is the concept of human-AI symbiosis combining computational efficiency with domain-specific human intuition \cite{mahmud2023study}. 
Facilitating seamless human-AI interactions allows humans to guide algorithms dynamically, ensuring that models align with user expectations \cite{saha2023human} even in the absence of specialization~\cite{choudhary2025human}. 
Especially for latent space inspections and modifications, human expertise is relevant to properly comprehend the active learning progression and its emerging challenges \cite{wei2022spaceediting}.


Challenges such as trust~\cite{paleja2024designs} and cognitive barriers~\cite{wolf2019evaluating} still delay the integration of HITL-based AI systems into traditional workflows, necessitating a compromise between explainability, interactivity, and usability to enhance human-AI collaboration.

\section{Designing for Integration of Human Intuition}

By synergistically infusing human expertise into ML pipelines, we can unlock unprecedented levels of model interpretability and performance, ultimately bridging the gap between artificial and human intelligence as we strive to replicate the nuanced complexities of human cognition \cite{gil2019towards}.
However, this integration requires careful consideration to balance human knowledge and intuition with the algorithmic nature of ML models \cite{sun2020evolution}. 
Unlike explicit domain knowledge, which is often systematic and well-documented, human intuition can be subjective, biased, and rooted in individual experiences \cite{wang2021interactive}. 
Misinterpreting intuition as knowledge or over-relying on it can lead to misguided decisions and suboptimal model behavior.

\subsection{Pitfalls of Human Intuition}

Every individual, whether an expert or novice in the field, brings unique experiences, beliefs, and perspectives, especially when designing and training complex deep learning models \cite{shrestha2021bias}. 
Such user-dependent biases, often unconscious, profoundly stem from the intricate nature of human decision-making, which often incorporates subjective interpretations, heuristics, perceptions, and personal experiences into the process. 
This unconsciousness can influence how data is processed, incorporated, and optimized, leading to inconsistent and potentially biased outcomes \cite{baer2017controlling}.
A prominent example can be observed in the training of large language models, where datasets represent the worldwide as well as local biases, imbalances, and user preferences that contribute to unintended outcomes, such as skewed representations or unfair predictions \cite{dhamala2021bold}.

Beyond those dataset issues, the direct insertion of human knowledge into the algorithmic training process introduces additional challenges. 
Overreliance on human knowledge, especially from single or small user groups, can result in aggressive interventions that compromise the model’s generalization capabilities, leading to overfitting to specific user preferences or beliefs \cite{chakraborty2021bias}. 
This problem becomes exacerbated when users lack complete knowledge of the system or its intricacies, leading to suboptimal guidance through intuition rather than knowledge.
However, the model integrates those inputs as ground truth, relying on the user's subjective biases and limitations.

Therefore, we aim to establish a clear distinction between guidance and manipulation. 
Human intuition should function as a subtle influence, offering directional cues without dictating the entire learning process. 
This differentiation is essential to prevent the model from becoming a mere extension of individual biases, which would undermine its objectivity and robustness. 
AI systems should leverage human intuition as a complementary force, to enhance their capabilities while maintaining the algorithmic integrity.

Finally, the varying strategies individuals employ when interacting with ML models or visualizations further underscore the pitfalls of human intuition \cite{monarch2021human}. 
Users may interpret and manipulate data differently, for instance by clustering, spreading, or reordering data points according to subjective criteria, potentially introducing inconsistencies into the training process. 
Addressing these challenges requires designing systems that feedback the human input with algorithmic checks such as validation performance metrics, ensuring that models remain guided but not dominated by human intuition.

\subsection{Training Guidance Through Human Intuition}
\begin{figure}[ht] 
    \centering
    \includegraphics[width=\textwidth]{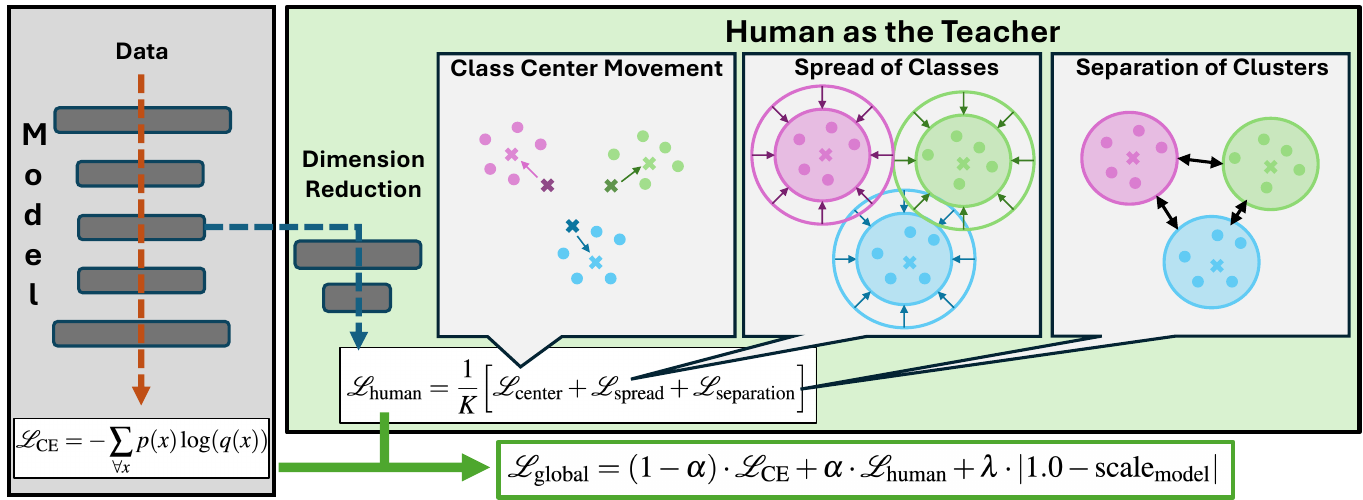} 
    \caption{Utilizing a weighted loss function through $\alpha$ to balance the classic cross-entropy with the human as the teacher. The human input is gathered from center movement, spread of classes and separation of clusters, which are finally normalized over the total number of pairwise comparisons $K$. Additionally the scale of the model is added in the global loss to regulate the loss function further. }
    \label{fig:kd_approach}
    \vspace{-0.5em}
\end{figure}

Including the human into model training requires a symbiotic learning paradigm where algorithmic optimization and human expertise interact synergistically. 
Our approach draws inspiration from the widely-used knowledge distillation technique, traditionally involving a teacher-student setup \cite{hinton2015distilling}. 
In typical scenarios, a large, pretrained teacher model transfers its knowledge to a smaller student model during training, using a shared loss function to improve the student’s performance. 
In this work, we replace the large model with a human expert, who acts as the teacher, guiding the model through intuitive adjustments and domain-specific insights.
The general framework behind \system{} can be found in \Cref{fig:kd_approach}, outlining the idea of human guidance instead of manipulation.

A core aspect of our framework is interactive visualization, which bridges human intuition and model training by allowing users to observe and refine latent space representations. Unlike non-deterministic techniques like t-SNE~\cite{van2008visualizing} or UMAP~\cite{mcinnes2018umap}, we use a deterministic transformation via fully connected layers to project high-dimensional latent data into a stable, interpretable 2D space.
This transformation layer is frozen after the first epoch, ensuring that changes in the visualization reflect model updates rather than mapping variations.

As depicted in \Cref{fig:kd_approach}, we measure the human-guided interventions and strategies based on three key metrics: the class center movement, considering the movement of whole class clusters and therefore shifting the center of the class; the spread of classes, representing the compactness change of each class such as moving all data points closer or further away from its designated center; and separation of clusters, measuring the distance of clear boundaries between individual class clusters. 
These modifications are incorporated into the training process through a weighted loss function. 
Rather than directly manipulating transformed data back into the original feature space, which usually requires a lossy and error-prone inverse of the dimension reduction, the human teacher’s adjustments are encoded within $\mathcal{L}_\text{human}$, a loss function that includes the center alignment, spread, and separation. 
Combined with the standard cross-entropy loss $\mathcal{L}_\text{CE}$, the global loss  $\mathcal{L}_\text{global}$ integrates both human guidance and algorithmic training, while the adjustable parameter $\alpha$ allows control for the level of human knowledge insertion.
As a guideline, we experimented $\alpha$ and selected 0.5 as the balance in order to maximize the human guidance while preserving the cross-entropy-based classification nature.
The additional $\lambda \cdot |1.0 - \text{scale}_\text{model}|$ term in the global loss ensures that the model maintains a consistent scale for features during training, preventing unintended distortions or penalization due to the range of the latent space visualizations and modifications.
Throughout our experiments, fixing $\lambda$ to 0.1 proved to regulate the strength of the model scale and stabilize the overall human guidance approach.

\subsection{The Latent Loop Tool (\system{})}
\label{sec:tool}
To effectively integrate the human as a teacher according to the framework from \Cref{fig:kd_approach}, we developed \system, an interactive tool designed to facilitate human-model collaboration. 
The tool enables users to inspect, analyze, and guide the latent space representations during training, ensuring a balance between learned model representations and domain-specific human knowledge according to $\mathcal{L}_\text{global}$.
We implemented an iterative and interactive process, in which the model is trained epoch-wise through the tool, allowing the user to pause the training to visualize the latent representation and feedback the latent adaptations for the next training iteration.
Noteworthy, the architecture of the tool is based on PyTorch \cite{NEURIPS2019_9015}, designed to allow individual, custom models, and datasets through two appropriate code interfaces to insert relevant data loaders and model architecture definitions.

\begin{figure}[ht] 
    \centering
    \includegraphics[width=0.8\textwidth]{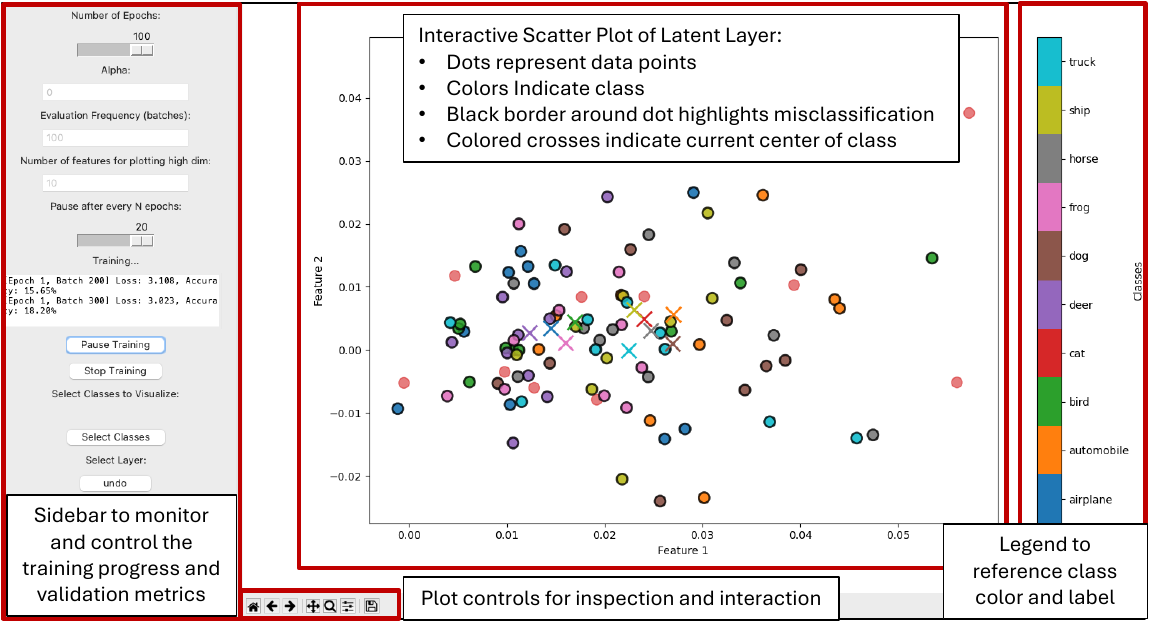} 
    \vspace{-0.5em}
    \caption{The user interface of \system{}; sidebar on the left to control the model training through the tool; main window obtaining the interactive scatter plots with relevant controls; a legend on the right to reference class labels with colors.}
    \label{fig:ui}
    \vspace{-0.5em}
\end{figure}

The tool’s UI comprises three main components as shown in \Cref{fig:ui}: (1) a control panel as a sidebar on the left, which provides functionalities such as training pause/resume, epoch adjustments, hyperparameter adjustment for the weighted loss function, and a text box printing the current validation metrics; (2) a scatter plot visualization as the main part of the window, displaying a two-dimensional scatter plot projection of the reduced latent space for intuitive inspection and guidance; and (3) a class reference legend on the right side, allowing users to map visual representations to corresponding class labels. 
The scatter plot visualization strategy has been selected as it represents a human-interpretable projection of the latent space, enabling intuitive interventions by moving points and clusters in a well-established drag-and-drop manner \cite{schneider2023visual}.
Apart from color-coding the individual classes, we add the class cluster centers as colored crosses and mark misclassifications with a black border around the dots.
As a main modification functionality, either individual dots can be moved or the whole cluster of a class can be moved by dragging the cluster center cross.

\begin{figure}[ht] 
    \centering
    \includegraphics[width=0.9\textwidth]{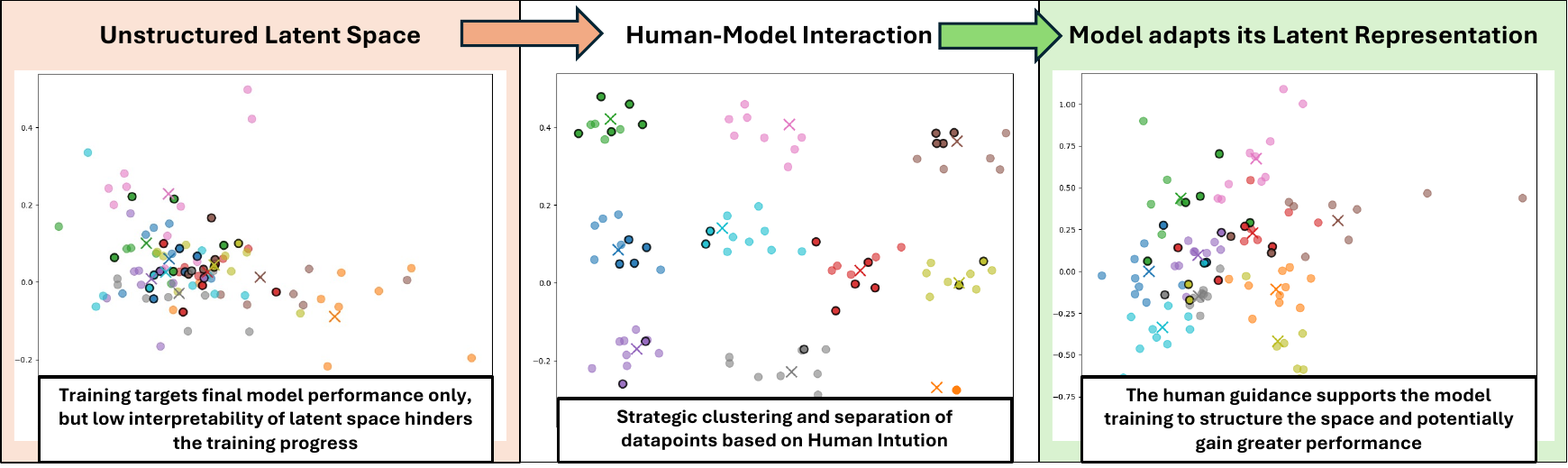} 
    \vspace{-0.5em}
    \caption{The extracted latent space of an exemplary training iteration. The model's latent space is initially unstructured and the model struggles to separate the classes properly, whereas after the insertion of human guidance, the model adapts its internal representation and classes can be distinguished with greater accuracy.}
    \label{fig:flowchart}
\end{figure}

As illustrated in Figure \ref{fig:flowchart}, the interaction loop consists of three key stages: (1) initial state, where the model generates an unstructured latent representation as its initial representation as part of the traditional training; (2) the human-model interaction, where users intuitively guide the model’s representation by clustering, spreading, or repositioning data points depending on the individual strategy; and (3) the adapted representation, where the model updates its internal structures to incorporate the human adjustments. 
This feedback loop allows human intuition to serve as a guiding force rather than a manipulative intervention, ensuring that the model does not merely overfit to user-imposed patterns but instead refines its latent space meaningfully.

\section{Evaluation}
We evaluate \system{} using a mixed-methods approach collecting qualitative feedback through observing tool usage and interviewing participants as well as encouraging them to voice their thoughts (think-aloud) during the experiment. 
Additionally, we administer established questionnaires on system usability (UMUX-Lite~\cite{lewis2013umux}) and perceived workload (NASA-TLX~\cite{hartDevelopmentNASATLXTask1988,hart2006nasa}) complemented by custom questions (\Cref{tab:custom_questions}) aimed to gauge the users' understanding of \system{} as well as its ability to support users in their task.
Finally, we compare the traditionally trained models with the models trained during the user study, to outline changes in classification performance in comparison with the applied interaction strategies.

\begin{table}[htb]
\footnotesize
	\caption{Custom questions, targeting the experience with the system. All rated on a visual analog scale (VAS) from 0 to 100; strongly disagree to strongly agree.}
	\label{tab:custom_questions}
	\centering
	\begin{tabular}{ll}
			\toprule
			\multicolumn{2}{l}{\textbf{Custom questions targeting the experience with the system.}}\\
			\midrule
			\textbf{Q1} & The system supported me in finding an optimal ML model.\\ 
            \textbf{Q2} & The system distracting me during my task.\\
            \textbf{Q3} & The system allowed me to understand the effectiveness of my model.\\
			\textbf{Q4} & The system biased me.\\
			\textbf{Q5} & I trust my model to perform.\\
			\textbf{Q6} & I fully understand the visualization of my model.\\
			\textbf{Q7} & The system made me understand the weaknesses of my model.\\
			\bottomrule
		\end{tabular}
        \vspace{-0.5em}
\end{table}

Our study consisted of two scenarios in which the participants were tasked to fine-tune a given prediction model. 
The two scenarios are based on the publicly available datasets CIFAR-10~\cite{krizhevsky2009learning} and PAMAP2~\cite{reiss2012introducing}. 
CIFAR-10 is an image dataset and contains low-resolution images across 10 classes, while PAMAP2 is a time-series dataset capturing human activities from wearable sensors across 13 classes. 
The model for CIFAR-10 consists of five convolutional layers with ReLU activation and dropout, interleaved with max-pooling layers, followed by four fully connected layers, including a final classification layer.
For PAMAP2, the model consists of three 1D convolutional layers with dropout, followed by global max and average pooling and a fully connected classification head.
The visualizations for both were generated based on the intermediate layer output after the convolutions before passing the data into the fully connected layers.
At this stage, the latent representation obtains the greatest insight while still not being affected by the final classification.
For both scenarios, participants were asked to use \system{} to find an optimal model to predict the classes. 
We did not set a model performance goal to achieve nor did we advise participants on possible strategies. 
Participants were only given a short introduction
into \system, highlighting its features and interaction options as well as explaining the latent space visualization as presented in \Cref{sec:tool}.

\subsection{Procedure}
After providing informed consent, we asked participants to provide demographics and their experience with ML models. A total of 14 participants (age: $\samplemean=27.1\,y$, $\samplesd=3.8\,y$; 5 female, 9 male) took part in our study. 
Participants reported an average experience with ML at $\samplemean=52.9$ ($\samplesd=28.8$, 0 to 100).
Subsequently, we introduced the participants to \system{} and randomly (counter-balanced) chose a starting scenario (either CIFAR-10 or PAMAP2).
In order to to prevent excessive experiment durations, we pretrained both models for 25 epochs.
With that, we cover two different scenarios, since CIFAR-10 still has great potential for improvement at this stage while PAMAP2 presents a challenge with less potential for improvement due to the complex nature of sensor data.
We asked participants to voice their thoughts during the interaction and tasked them to improve the performance of the given model. 
We did not put any interaction constraints on this step and allowed participants to freely explore \system{} and strategies they saw fit to improve the model performance. 
However, to receive comparable results, we fixed four interaction points in the training, respectively at epochs 25, 30, 35, and 40.
After each training iteration (initiated by the participants at will), the updated model performance was displayed and the respective latent space visualization was updated. 
After each full pass of a scenario, we administered the UMUX-Lite, NASA-TLX, and our custom questions. 
Participants subsequently were tasked with the second, remaining scenario. 
At the end of the study, we interviewed the participants on their impressions of the system. 
The study duration did not exceed 60\,min and was approved by the Ethics Team of the German Research Center for Artificial Intelligence.


\subsection{Results}
\label{sec:results}
We analyzed the administered questionnaires and the recorded think-aloud protocols including the interviews to evaluate the user experience of \system. We particularly focused on distilling the strategies to improve the model employed by the participants. A comparison of the model performance using \system{} and baseline training without human intervention provides insight into the effectiveness of our approach.

\subsubsection{Model Performance}
\label{sec:model_performance}
\begin{figure}
     \centering
     \begin{subfigure}{0.49\textwidth}
         \centering
         \includegraphics[width=\linewidth]{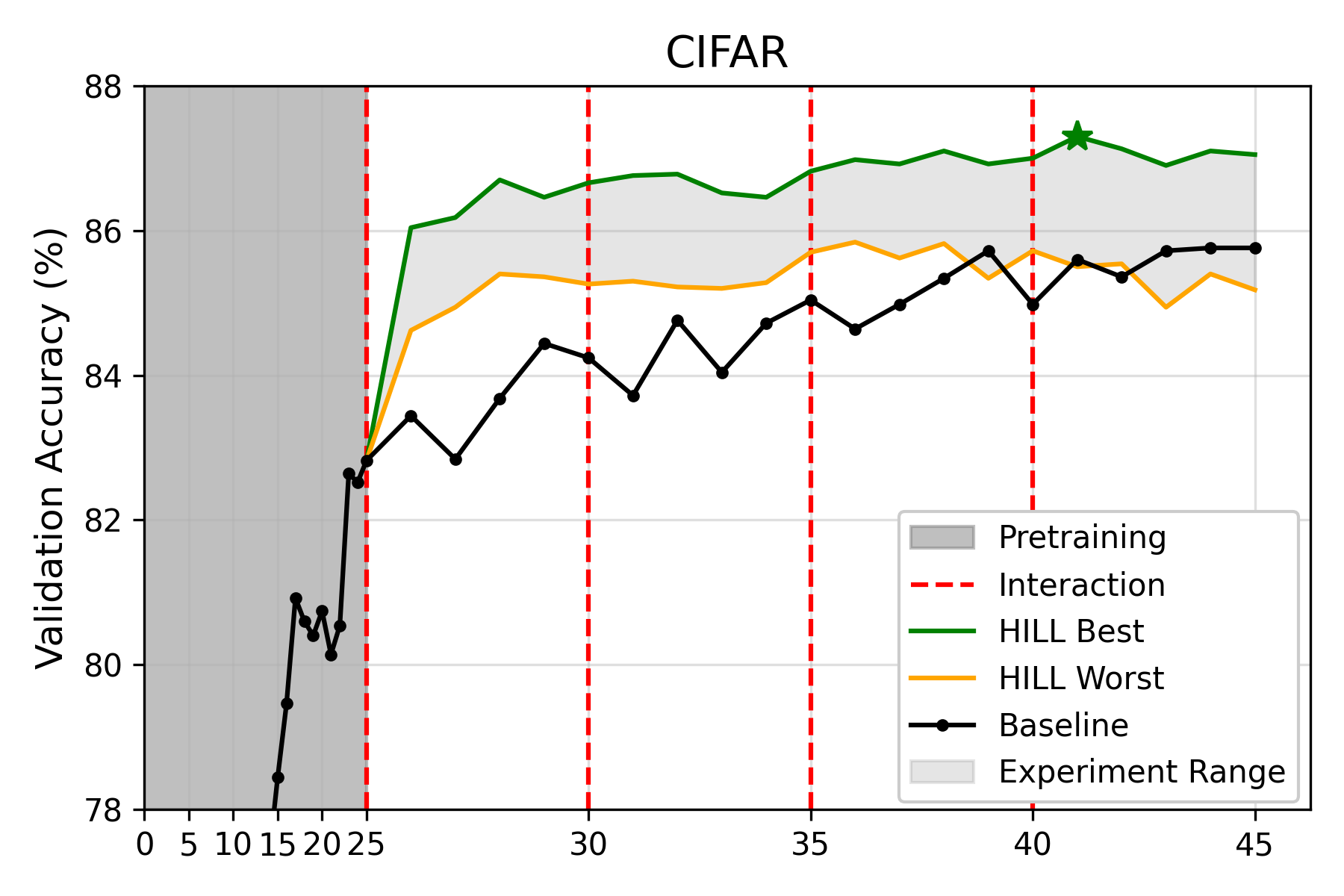}
         \label{fig:cifar_perf}
     \end{subfigure}
     \hfill
     \begin{subfigure}{0.49\textwidth}
         \centering
         \includegraphics[width=\linewidth]{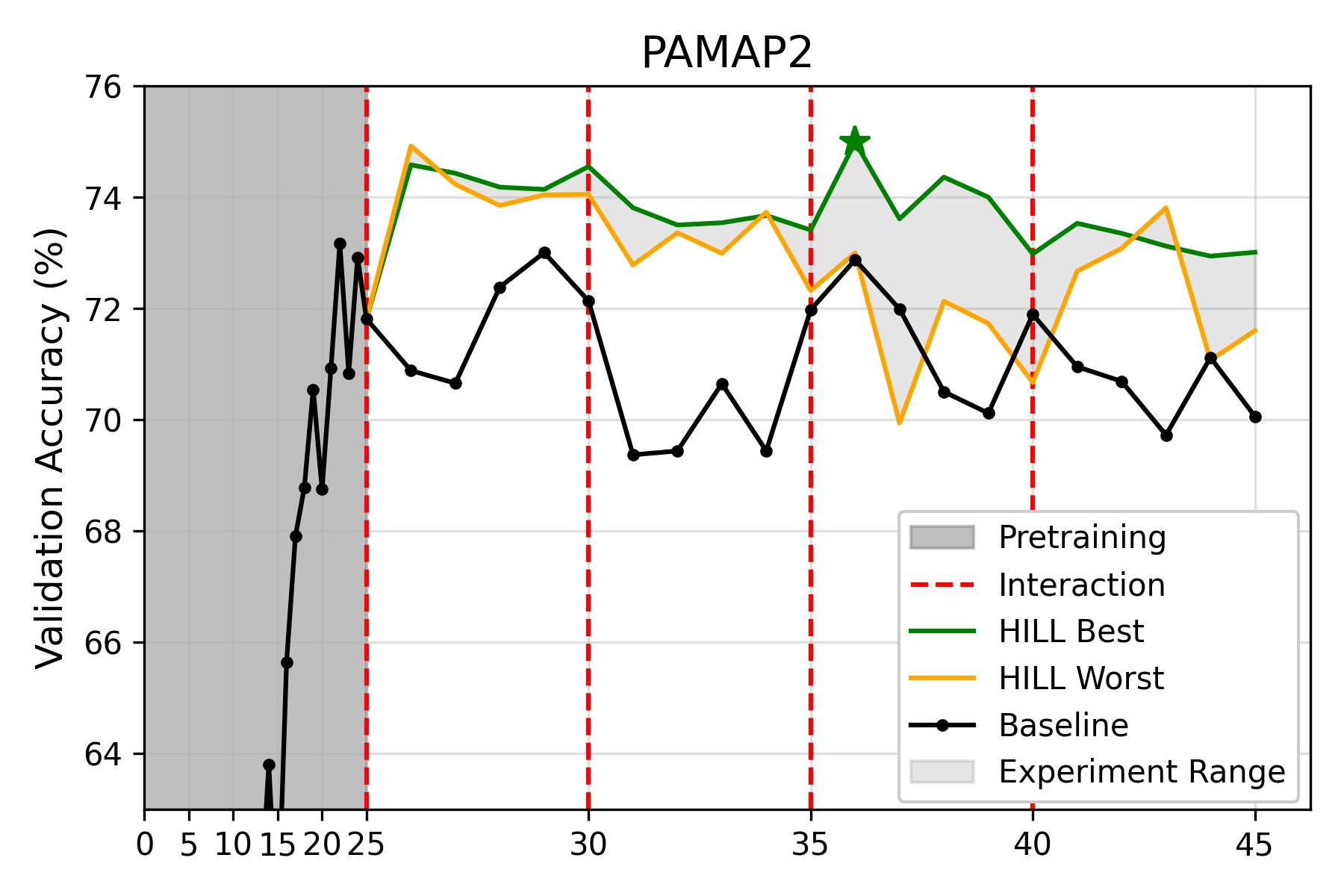}
         \label{fig:pamap_perf}
     \end{subfigure}
     \vspace{-2em}
     \caption{The model performance evaluation of \system{} compared to the traditional training as baseline; Dark grey represents the pretraining phase; red dashed lines human interaction points; the light grey area represents the range of best (green) and worst (orange) participant utilizing \system{} during the user study.}
     \label{fig:performance}
     \vspace{-0.5em}
\end{figure}

We present the validation accuracy for the two scenarios in Figure~\ref{fig:performance}, comparing the traditional training (baseline) with \system{}. We visualize the range of participants' experiments of \system{}, including the best-performing experiment in \textit{green} and the worst in \textit{orange}.
The evaluation was conducted on the unseen validation dataset, the same as during training, to ensure that the model does not overfit the test data.

Across both datasets, the first interaction had the most significant impact on performance. 
This can be attributed to the user’s initial structuring of the latent space, which often resulted in a great boost of accuracy. 
Even the worst-performing \system{} results were still capable of outperforming the baseline, demonstrating the robustness of the approach.
Notably, the best-performing \system{} runs significantly surpassed the baseline, providing strong evidence that human intuition can effectively enhance training. 
For CIFAR, the accuracy increased by 1.6\% points to 87.3\%, while for PAMAP2, the improvement was around 2.2\% points to 75\%. 
Additionally, \system{} led to faster convergence, reducing the computational resources required to achieve satisfactory classification performance.
In the case of PAMAP2, the worst \system{} performance can be traced back to a participant who altered their strategy across interaction points. 
This change resulted in greater fluctuations in accuracy, particularly during the last two interactions.

\subsubsection{Questionnaires}
\label{sec:questionnaires}
\begin{figure}[ht] 
    \centering
    \includegraphics[width=0.7\textwidth]{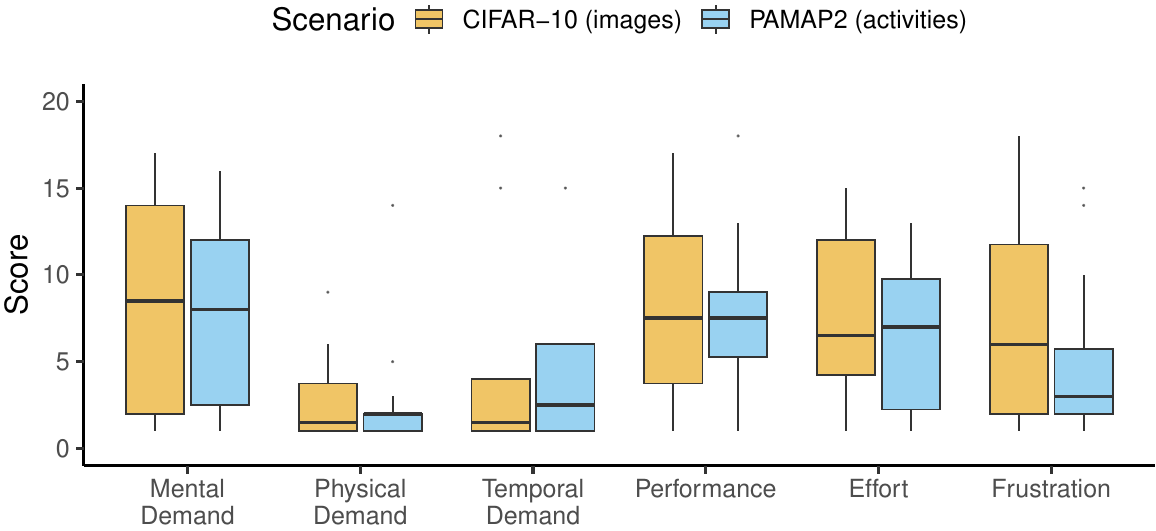} 
    \caption{Scores for NASA-TLX subscales (0 to 20) given our two scenarios (\scenarioA, \scenarioB).}
    \label{fig:tlx}
    \vspace{-0.5em}
\end{figure}

We calculated the SUS-parity score ($max=100$) from the collected UMUX-Lite responses~\cite{lewis2015investigating}. Our system showed good usability~\cite{bangor2009determining} with $\samplemean=72.4$ ($\samplesd=9.23$). Likewise, the total NASA-TLX ($max=120$) indicated a low workload for participants with $\samplemean=36.1$ ($\samplesd=18.2$). \Cref{fig:tlx} depicts the individual subscales ($max=20$) given our two scenarios. Physical and temporal demand is very low, while other subscales score slightly higher with bigger variance. In particular, frustration was lower for \scenarioB{}.

\begin{figure}[ht] 
    \centering
    \includegraphics[width=0.7\textwidth]{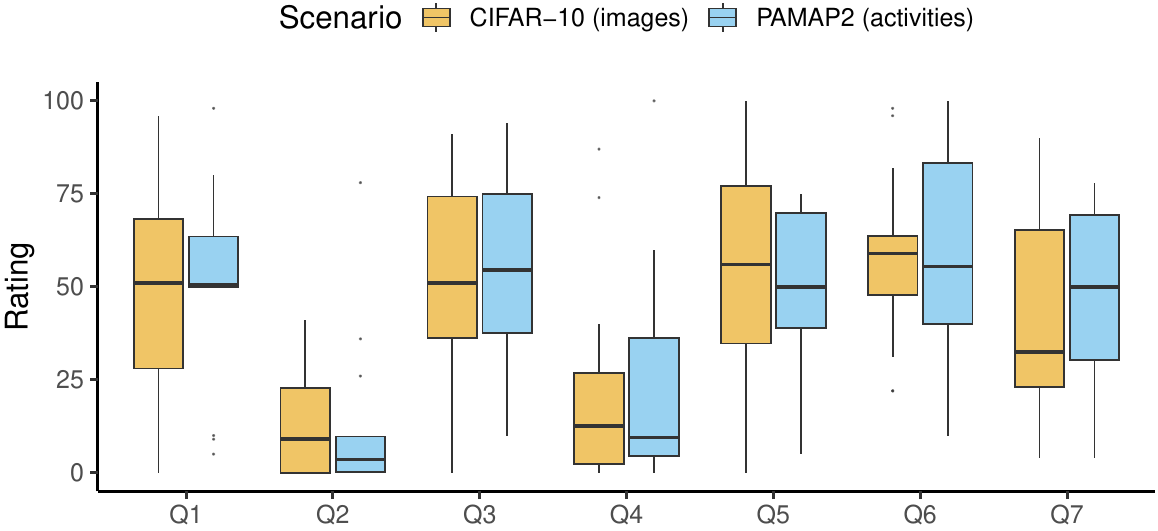} 
    \caption{Ratings for our custom questions (\Cref{tab:custom_questions}) given our two scenarios (\scenarioA, \scenarioB).}
    \label{fig:custom}
    \vspace{-1em}
\end{figure}

The results of our custom questions (\Cref{tab:custom_questions}) are visualized in \Cref{fig:custom}. For Q2 ("The system distracting me during my task.") and Q4 ("The system biased me.") we recorded low ratings from the participants across both scenarios indicating that the system was neither distracting the user ($\samplemean=12.6$, $\samplesd=18.1$) nor biased their decision making ($\samplemean=23.1$, $\samplesd=28.1$). For all other questions, we recorded split ratings, associated with whether participants were successful in improving the model with their strategies. This affected their model beliefs, such as its effectiveness (Q3), its weaknesses (Q7), and their trust in the model (Q5). Notably these beliefs also translated to an understanding of the visualization (Q6) and whether the system was able to support them in finding an optimal model (Q1). In particular, for Q1, we found that scores varied less for the \scenarioB{} scenario, also linking to its lower frustration score (NASA-TLX).

\subsubsection{Think-aloud and Interviews}
\label{sec:thinkaloud}
We transcribed the audio files (10 hours 30 minutes in total) collected from each participant using Whisper \cite{radford2022whisper}. 
To analyze the interviews, we used the approach by Blandford et al. \cite{blandford2016qualitative}.
Three researchers coded an initial 20\% of the data and agreed on an initial coding tree. The remaining data was split between the researchers.
From a final discussion, the following themes surfaced: \textit{Strategies}, \textit{Human-Model-Interaction}, and \textit{User Interface}.
Since \textit{Strategies} was the main topic for most participants during the study we subdivided that topic into six subtopics.

\paragraph{Strategies}
Most participants agreed on moving data points closer to their cluster center in order to remove misclassified outliers, therefore \textbf{Increasing Cluster Compactness}.
\begin{quoting}\noindent
\textit{
"Now my strategy is to cluster them near to the exact center." (P13)
}\end{quoting}
Another commonly applied strategy involved \textbf{Maximizing Cluster Distance}, where participants moved whole clusters further apart from each other, in an effort to help the model distinguish and separate them properly.
\begin{quoting}\noindent
\textit{
"Let's get the airplane away from the bird." (P14)
}\end{quoting}
Contrary to moving clusters apart, some participants applied the \textbf{Merge Similar Classes} strategy where they decided to move classes connected on a high abstraction level closer.
\begin{quoting}\noindent
\textit{"(...) feature one, you have animals, and then kind of sort them. And on the other side, like, cars, automobiles, like transportation things." (P14)
}\end{quoting}
In cases where participants were shown dense, well-separated clusters with a clear distinction from others, they commonly employed the \textbf{Keep Arrangement} strategy, only moving points from classes where this was not the case.
\begin{quoting}\noindent
\textit{"I think the automobile is great. So I'll not move this at all." (P4)
}\end{quoting}
When the selected strategy did not produce satisfying improvements, a few participants \textbf{Changed Strategy} and decided to continue training by inverting their previous approach.
\begin{quoting}\noindent
\textit{"I would like to try something completely different. Before I purposefully tried to separate classes that are similar. But what happens if I do the exact opposite?" (P6)
}\end{quoting}

\paragraph{Human-Model-Interaction.}
Apart from diverse strategies, most of the participants gave qualitative feedback about their impressions working directly with the model during training. Most feedback was positive, showing that \system{} helps understanding the inner workings of the trained models.
\begin{quoting}\noindent
\textit{"It's not so much a black box anymore" (P2)
}\end{quoting}
Yet, some participants wondered about the lack of influence they had on the training.
\begin{quoting}\noindent
\textit{"I wanted to see how the tool reacts, if it helps with training at all or not.
So I tried a few different things and I think for the first task this worked somewhat, for the second task really not." (P6)
}\end{quoting}

\paragraph{User Interface}
Overall, participants appreciated features and layout of the interface.
However, some required more feedback if their actions had the desired impact.
\begin{quoting}\noindent
\textit{"The system is easy to use, but I'm obviously doing something wrong." (P2)
}\end{quoting}

\section{Discussion}
A key challenge in the field of HITL ML is that users bring their own preconceived notions into the interaction, which can subtly shape both their engagement with the system and the resulting model behavior. 
As indicated by our interviews (\Cref{sec:thinkaloud}), participants often approached latent space modifications based on their own mental models of the data, sometimes \textbf{reinforcing subjective structures rather than optimizing for pure algorithmic generalization}. 
This phenomenon links to developer blindness, where those designing or interacting with such systems impose their own interpretations into the model, overlooking alternative perspectives or emergent patterns \cite{johansen2020studying}.

Unlike related approaches that modify training data directly (cf.~\Cref{sec:RL}), \system{} operates within the latent space, only including it in the loss function, hence, making it less susceptible to overfitting caused by human biases as evident from our analysis on model performance (\Cref{sec:model_performance}). 
However, guidance should not turn into manipulation, since allowing too much user influence can lead to the model conforming to subjective strategies rather than generalizable improvements \cite{daee2018user}.
Especially when evaluating the model performance, users need to ensure to exclude their own beliefs and intuitions to properly \textbf{balance human intuition with algorithmic training}.

Human users interpret data differently than algorithms, often applying concepts based on their knowledge about the classes, such as grouping or separating “birds” and “planes” since they are correlated with the sky.
This can enrich training by \textbf{introducing structured insights from human cognition}, which purely data-driven approaches overlook. 
Our results show that human-induced adjustments improved model performance without causing overfitting (\Cref{sec:model_performance}), suggesting that human guidance can highlight relevant decision boundaries that standard optimization fails to capture. 
Future work should explore adaptive strategies where the model learns which human interventions are beneficial and adjusts its reliance accordingly.

A recurring topic in user feedback was the need for more \textbf{transparent algorithmic responses to the user}, apparent in our questionnaires (\Cref{sec:questionnaires}) and interviews (\Cref{sec:thinkaloud}). 
Some participants struggled to gauge whether their interactions were meaningful, which aligns with findings from our think-aloud analysis. 
Addressing this requires more explicit feedback mechanisms beyond pure performance metrics, such as visualizing the differences between latent spaces before and after intervention. 
Providing explanatory cues, for instance highlighting which modifications or classes contributed to accuracy improvements, or even utilizing support from Large Language Models~\cite{krupp2024llm}, could further bridge the gap between human intuition and model behavior.

Traditional training optimization techniques focus on minimizing a loss function defined over the training dataset, which encapsulates the differences between predicted and actual outcomes within that specific context. While these methods are effective for achieving reasonable performance on the given data without human intervention, they operate within the confines of the information provided by the dataset.
Human intuition, on the other hand, draws upon a vast reservoir of world knowledge, experiences, and contextual understanding that extends far beyond the boundaries of any single dataset. This intuitive knowledge includes common sense reasoning and domain-specific expertise. As such, \textbf{human intuition has the ability to recognize patterns or relationships that may not be explicitly represented in the data} but are crucial for real-world applications.

\section{Conclusion}
In this work, we introduced "Human in the Latent Loop" (\system), facilitating a novel human-AI interaction paradigm that interactively infuses human intuition into model training through latent space representations. \system{} demonstrated that human interventions improve model performance and convergence while maintaining generalization. Our evaluation revealed diverse interaction strategies, highlighting both the strengths as well as the risks of potential human biases. 
By integrating human intuition into the model training process, particularly through interactive methods such as manipulating latent space clusters, HILL can effectively bridge the gap between data-driven optimization and human-centric understanding. 
This approach enables the model to benefit from insights that traditional methods might overlook, potentially leading to solutions that not only perform well on the dataset but also make sense in broader, real-world contexts.

\section*{Acknowledgement}
This work is supported by the European Union’s Horizon Europe research and innovation program (HORIZON-CL4-2021-HUMAN-01) through the "SustainML" project (grant agreement No 101070408).

\bibliographystyle{plain}
\bibliography{bibliography}
\end{document}